\documentclass[letterpaper]{article} 
\usepackage{aaai2026}  
\usepackage{times}  
\usepackage{helvet}  
\usepackage{courier}  
\usepackage[hyphens]{url}  
\usepackage{graphicx} 
\usepackage{enumitem}
\usepackage{natbib}  
\usepackage{caption} 
\usepackage{algorithm}
\usepackage{algorithmic}

\usepackage{tabularx}
\usepackage{xcolor}
\usepackage{colortbl}
\usepackage{booktabs}
\usepackage{multirow}
\usepackage{multicol}
\usepackage{subfigure}
\usepackage{enumitem}

\newtheorem{definition}{Definition}

\usepackage{bm}
\usepackage{amsmath}
\usepackage{amssymb}
\usepackage{mathrsfs}

\usepackage{adjustbox}

\usepackage{newfloat}
\usepackage{listings}
\DeclareCaptionStyle{ruled}{labelfont=normalfont,labelsep=colon,strut=off} 
\floatstyle{ruled}
\newfloat{listing}{tb}{lst}{}
\floatname{listing}{Listing}
\title{Meta Dynamic Graph for Traffic Flow Prediction}
\author{
    Yiqing Zou\textsuperscript{\rm 1}, Hanning Yuan\textsuperscript{\rm 1}, Qianyu Yang\textsuperscript{\rm 1}, Ziqiang Yuan\textsuperscript{\rm 1}, Shuliang Wang\textsuperscript{\rm 1}, Sijie Ruan\textsuperscript{\rm 1}\thanks{Sijie Ruan is the corresponding author.}\\
}
\affiliations{
    \textsuperscript{\rm 1}Beijing Institute of Technology, Beijing, China\\

    \{zouyiqing, yhn6, yangqy, ziqiangy, slwang2011, sjruan\}@bit.edu.cn\\
}

\newcommand{\sstitle}[1]{\noindent{\bf #1\/.}}

\usepackage{bibentry}

\begin{document}

\maketitle

\begin{abstract}
Traffic flow prediction is a typical spatio-temporal prediction problem and has a wide range of applications. The core challenge lies in modeling the underlying complex spatio-temporal dependencies. Various methods have been proposed, and recent studies show that the modeling of dynamics is useful to meet the core challenge. While handling spatial dependencies and temporal dependencies using separate base model structures may hinder the modeling of spatio-temporal correlations, the modeling of dynamics can bridge this gap. Incorporating spatio-temporal heterogeneity also advances the main goal, since it can extend the parameter space and allow more flexibility. Despite these advances, two limitations persist: 1) the modeling of dynamics is often limited to the dynamics of spatial topology (e.g., adjacency matrix changes), which, however, can be extended to a broader scope; 2) the modeling of heterogeneity is often separated for spatial and temporal dimensions, but this gap can also be bridged by the modeling of dynamics. To address the above limitations, we propose a novel framework for traffic prediction, called \underline{Meta} \underline{D}ynamic \underline{G}raph (MetaDG). MetaDG leverages dynamic graph structures of node representations to explicitly model spatio-temporal dynamics. This generates both dynamic adjacency matrices and meta-parameters, extending dynamic modeling beyond topology while unifying the capture of spatio-temporal heterogeneity into a single dimension. Extensive experiments on four real-world datasets validate the effectiveness of MetaDG.
\end{abstract}

\begin{links}
    \link{Code}{https://github.com/zouyiqing-221/MetaDG}
\end{links}

\section{Introduction}

The advances of spatio-temporal data collection technology have made the study of spatio-temporal data increasingly prevalent. When modeling spatio-temporal data, it is essential to take into account spatial, temporal, and spatio-temporal correlations simultaneously. The properties of the temporal and spatial dimensions have many differences, and modeling the interaction properties of these two dimensions is even more difficult. Existing works, such as STGCN \cite{stgcn} and GWNet \cite{gwn}, etc., are based on the combination of a temporal model and a spatial model, which separately capture temporal and spatial dependencies \cite{stfgnn}. The separation of the base model makes it difficult to capture complex spatio-temporal dependencies. For convenience, we define modeling spatial and temporal dimensions separately as \textbf{ST-isolated}.

Further research shows that considering dynamics may be an effective way to put these two dimensions together and can capture cross-dimensional interactions in a more explicit manner. The way that they take dynamics into consideration is based on the observation that information propagation happens not only in spatial dimension, but also in temporal dimension. For instance, STSGCN \cite{stsgcn} tried to synchronously capture propagations on spatial, temporal, and spatio-temporal dimensions, while PDFormer \cite{pdformer} tried to capture propagation delays between spatio-temporal nodes. A more direct way to capture the dynamics is given by DGCRN \cite{dgcrn}, which generates a dynamic adjacency matrix for each time step in the time sequence. These methods have incorporated the modeling of dynamics into the model, yet they limit the usage of dynamics within affecting spatial topology \cite{2sagcn, astgcn, mtgnn, dgcrn}. Indeed, focusing on spatial topology and ignoring latent semantics will strongly limit the performance \cite{stgode}. Thus, we suggest that the modeling and usage of dynamics can be more general and influential than we used to know. That is, considering dynamics can push \textbf{ST-isolated} towards \textbf{ST-unification}. Since the modeling of dynamics is often limited to generating dynamic adjacency matrices, we attempt to extend the usage of dynamics to a broader scale.

Incorporating spatio-temporal heterogeneities may also enhance the modeling of spatio-temporal dependencies~\cite{ruan2025spatial}. AGCRN \cite{agcrn}, MegaCRN \cite{megacrn}, and HimNet \cite{himnet} attempt to generate adaptive node representations and further generate adaptive adjacency matrices and meta-parameters so that they can model spatio-temporal heterogeneities. Incorporating heterogeneities has been tested to be effective, but with an ST-isolated base model structure, the modeling of spatio-temporal heterogeneities also faces the same problem of ST-isolation. As we have seen that modeling dynamics can draw base model from ST-isolated towards ST-unification, it is reasonable to incorporate dynamics into the modeling of spatio-temporal heterogeneities.

To address the above limitations in the mentioned way, we propose a novel framework for traffic prediction, called \underline{Meta} \underline{D}ynamic \underline{G}raph (MetaDG). MetaDG uses GCRU \cite{agcrn, dgcrn, himnet} as the base model structure of both encoder and decoder, and leverages dynamic graph structures of node representations to explicitly model spatio-temporal dynamics. Specifically, MetaDG first generates raw dynamic node embedding for each time step according to the Dynamic Node Generation module. In order to enhance the dynamic node embedding generated in each time step, we further design a Spatio-Temporal Correlation Enhancement module, so that each node can extract information from historical node representations and properly smooth out differences across time steps. Since the reliability of message-passing is of critical importance in GNN-based models \cite{ungsl} and slight errors may accumulate according to the recurrent nature of RNN-based models, we further propose a Dynamic Graph Qualification module to refine the adjacency matrix by measuring the qualification of information propagations. These components eventually give out the Meta Dynamic Graph Convolutional Recurrent Unit, where we generate meta-parameters, raw adjacency matrix, and edge-weight adjustment matrix for graph convolution at each time step.

We summarize our contributions as follows:
\begin{itemize}[leftmargin=*]
    \item We generate dynamic node embedding and enhance the node representation by spatio-temporal correlations for each time step. By modeling the dynamic graph structure of the spatio-temporal nodes, we can bridge the gap between the two dimensions and push the \textbf{ST-isolated} base model towards \textbf{ST-unification}.
    \item We illustrate the importance of considering the reliability of message-passing, and therefore propose to refine the adjacency matrix according to the qualification of information propogation by generating an edge-weight adjustment matrix.
    \item We use enhanced dynamic node representation and edge-weight adjustment matrix to generate meta-parameters and adjacency matrix for each time step. This allows us to incorporate both dynamics and heterogeneities. The usage of dynamics has been extended to a broader scope, and spatio-temporal heterogeneities have been modeled in an ST-unifying manner.
    \item Extensive experiments are conducted on four real-world datasets, which demonstrate the effectiveness of modeling dynamics and heterogeneities simultaneously.
\end{itemize}

\section{Preliminaries}\label{sec:prelim}

\begin{definition}[Road Network]
    A road network is a directed graph, denoted by $\mathcal{G}=(\mathcal{V}, \mathcal{E}, A)$. Here, $\mathcal{V} = v_1, ..., v_N$ denotes a set of $N = \vert \mathcal{V} \vert$ nodes representing different locations on the road network; $\mathcal{E} \subseteq \mathcal{V} \times \mathcal{V}$ is a set of edges; $A \in \mathbb{R}^{N \times N}$ is the adjacency matrix representing the spatial topology of the nodes. Note that we do not use a predefined adjacency matrix in our model.
\end{definition}

\begin{definition}[Traffic Flow]
    Let $\boldsymbol{X}_t \in \mathbb{R}^N$ denote the traffic flow of all $N$ nodes at time step $t$.
\end{definition}

\noindent \textbf{Problem Formulation.}
\textit{Traffic flow prediction aims to predict the traffic flow of a traffic system in the future period based on the observation of the historical period. Formally, given the road network $\mathcal{G}=(\mathcal{V}, \mathcal{E}, A)$ and the historical traffic flow of past $T$ time steps $[\boldsymbol{X}_{t-T+1}, ..., \boldsymbol{X}_t]$, learn the map $f$ which predicts the future traffic flow of $T'$ time steps $[\boldsymbol{X}_{t+1}, ..., \boldsymbol{X}_{t+T'}]$: 
\begin{equation}
    f: [\boldsymbol{X}_{t-T+1}, ..., \boldsymbol{X}_t; \mathcal{G}] \to [\boldsymbol{X}_{t+1}, ..., \boldsymbol{X}_{t+T'}].
\end{equation}
}

\section{Methodology}\label{sec:method}

\begin{figure*}[t]
\centering
\includegraphics[width=0.98\textwidth]{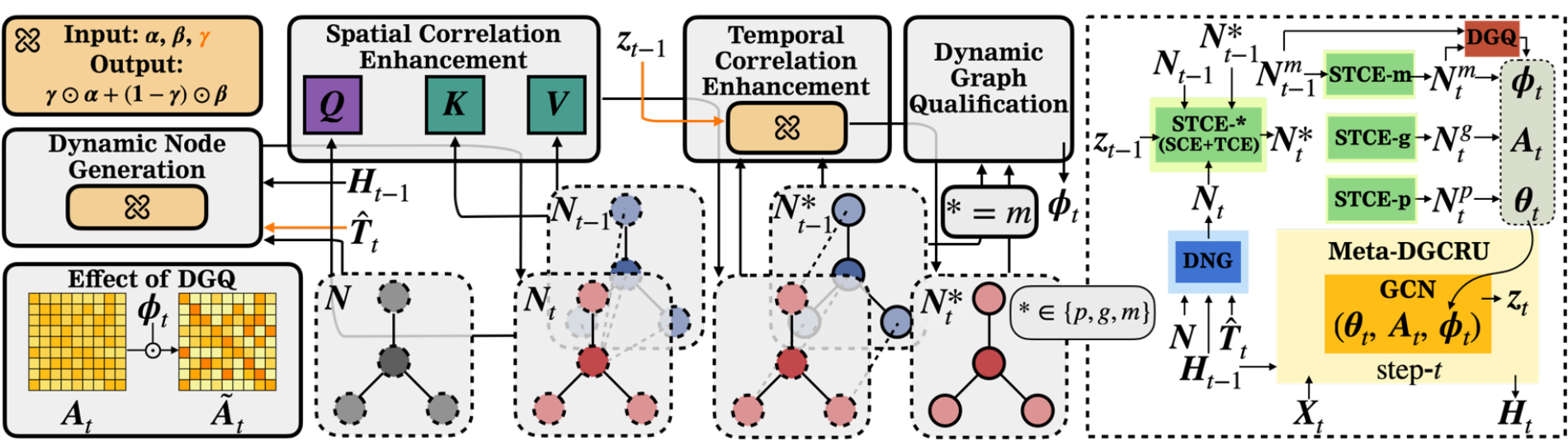}
\caption{Framework of MetaDG.}
\label{fig:metadg-framework}
\end{figure*}

\subsection{Overview}

We propose a novel framework, Meta Dynamic Graph (MetaDG), for traffic flow prediction, as shown in Figure \ref{fig:metadg-framework}. We use Graph Convolutional Recurrent Unit (GCRU) as the basic structure for the encoder and decoder, which is a combination of Gated Recurrent Unit (GRU) and Graph Convolutional Neural Network (GCN), since it is a sequence-to-sequence architecture often used for spatio-temporal predictions \cite{stgcn}. Compared with the standard GCRU, MetaDG uses dynamically generated adjacency matrix and meta-parameters at each time step. Specifically, at each time step, MetaDG generates raw dynamic node embedding. By learning the dynamic graph structure of all nodes, MetaDG models the structure of meta-parameters and the graph structure of information propagation. Hence, the thoughts of dynamics and heterogeneities have been incorporated into the framework of spatio-temporal prediction effectively and simultaneously. MetaDG has 3 main components: 1) Dynamic Node Generation (DNG) module, which can generate raw dynamic node embedding for each time step; 2) Spatio-Temporal Correlation Enhancement (STCE) module, which will enhance the raw dynamic node embedding based on spatio-temporal correlations across time steps, and will be further used to generate adjacency matrix and meta-parameters; 3) Dynamic Graph Qualification (DGQ) module, which will generate edge-weight adjustment matrix by measuring the qualification of information propagations on edges, and will be further used to refine the structure of the dynamically generated raw adjacency matrix. These modules will eventually bring us the dynamic adjacency matrix and meta-parameters used in the calculation of graph convolution, which construct the MetaDG Convolutional Recurrent Unit (Meta-DGCRU) for each time step.

\subsection{Dynamic Node Generation}

In this subsection, we propose the dynamic node generation (DNG) module that generates raw dynamic node embedding, and will be further enhanced in the following module. Urban traffic conditions are complex, and spatio-temporal correlations are highly dynamic \cite{dgcrn}. The dynamic property comes from highly variable real-time traffic conditions. Dynamics is exhibited not only in real spatial topology, as represented by current adjacency matrix, but also in traffic conditions propagated between nodes with time differences. 

To achieve enough flexibility, we do not use a predefined adjacency matrix, but use a learnable static node embedding to calculate the adjacency matrix based on inner product. Specifically, the static node embedding is denoted as $\boldsymbol{N} \in \mathbb{R}^{N \times d_s}$, where $N$ is the number of nodes, and $d_s$ is the spatial embedding dimension. Based on the static node embedding, for each time step $t$, we use current time embedding $\boldsymbol{T}_t \in \mathbb{R}^{B \times d_t}$ and previous hidden embedding $\boldsymbol{H}_{t-1} \in \mathbb{R}^{B \times N \times d_H}$ to generate current dynamic node embedding $\boldsymbol{N}_t$. Here, $B$ denotes the batch size, $d_t$ denotes the temporal embedding dimension, and $d_H$ denotes the hidden state dimension. Each $\boldsymbol{T}_t$ combines the embedding of time-of-day and day-of-week \cite{pdformer}, thus $d_t = d_{tod} + d_{dow}$. We further enhance $\boldsymbol{T}_t$ to be $\hat{\boldsymbol{T}_t} \in \mathbb{R}^{2d_t}$, 
to discriminate the same timestamp that appears in different time steps. That is, $\hat{\boldsymbol{T}_t} = $ $[\boldsymbol{T}_1 \Vert \boldsymbol{T}_t]$ for encoder, and $\hat{\boldsymbol{T}_t} = [\boldsymbol{T}_{-1} \Vert \boldsymbol{T}_t]$ for decoder.

To generate dynamic node embedding $\boldsymbol{N}_{t} \in \mathbb{R}^{B \times N \times d_s}$, use enhanced time embedding $\hat{\boldsymbol{T}_t}$ to obtain a time-based dynamic gate $\boldsymbol{\gamma}_t$, which will be used to fuse static node embedding $\boldsymbol{N}$ and current hidden state $\boldsymbol{H}_{t-1}$:
\begin{equation}
    \boldsymbol{N}_{t} = \boldsymbol{\gamma}_t \odot \boldsymbol{N} + (\boldsymbol{1} - \boldsymbol{\gamma}_t) \odot \hat{\boldsymbol{H}}_{t-1},
\end{equation}
where
\begin{equation}
    \hat{\boldsymbol{H}}_{t-1} = \mathrm{FC}_H({\boldsymbol{H}}_{t-1}),
\end{equation}
\begin{equation}
    \boldsymbol{\gamma}_{t} = \mathrm{sigmoid}(\hat{\boldsymbol{T}_t} \boldsymbol{\Gamma}).
    \label{eq:dynamic-gate}
\end{equation}
Here, $\odot$ denotes Hadamard product, $\mathrm{FC}_H(\cdot)$ is a map from $d_H$ dimensions to $d_s$ dimensions, $\boldsymbol{\Gamma} \in \mathbb{R}^{2d_t \times d_s}$ is a time embedding pool used to generate $\boldsymbol{\gamma}_t \in \mathbb{R}^{B \times d_s}$. The time-based dynamic gate $\boldsymbol{\gamma}_t$ allows the model to decide the strength of dynamics for every dimension at each time step. If $\boldsymbol{\gamma}_t$ is low, consider more about $\hat{\boldsymbol{H}}_{t-1}$ when generating $\boldsymbol{N}_{t}$. Consider more about $\boldsymbol{N}$ otherwise. That is, low $\boldsymbol{\gamma}_t$ indicates high flexibility.

\subsection{Spatio-Temporal Correlation Enhancement}
In this subsection, we enhance the raw node embedding $\boldsymbol{N}_{t}$ based on spatio-temporal correlations across time steps.

\subsubsection{Spatial Correlation Enhancement (SCE).}

Given raw dynamic node embedding $\boldsymbol{N}_{t}$, we first refine this representation based on spatial correlations. Note that $\boldsymbol{N}_{t}$ is not yet directly correlated with $\boldsymbol{N}_{t-1}$, so that the model may face learning difficulties due to drastic changes in dynamic node representations, and useful historical information might be lost due to frequent fluctuations. Thus, to integrate node representations from the previous time step with those of the current time step, thereby obtain historically enhanced dynamic node representation, we adopt cross-attention mechanism \cite{vaswani2017attention} to incorporate critical global historical information into the current dynamic node representation. 

Specifically, use $\boldsymbol{N}_{t}$ to obtain historical information from $\boldsymbol{N}_{t-1}$. That is, generate query $\boldsymbol{Q}_t$ from $\boldsymbol{N}_{t}$, and generate key $\boldsymbol{K}_t$ and value $\boldsymbol{V}_t$ from $\boldsymbol{N}_{t-1}$:
\begin{equation}
\boldsymbol{Q}_t = \mathrm{FC}_Q(\boldsymbol{N}_{t}),
\boldsymbol{K}_t = \mathrm{FC}_K(\boldsymbol{N}_{t-1}),
\boldsymbol{V}_t = \mathrm{FC}_V(\boldsymbol{N}_{t-1}),
\end{equation}
where $\boldsymbol{Q}_t, \boldsymbol{K}_t, \boldsymbol{V}_t \in \mathcal{R}^{B \times N \times d'}$; $\mathrm{FC}_Q(\cdot)$, $\mathrm{FC}_K(\cdot)$, and $\mathrm{FC}_V(\cdot)$ are fully connected layers mapping from $d_s$ dimensions to $d'$ dimensions. Use $\boldsymbol{Q}_t$, $\boldsymbol{K}_t$, and $\boldsymbol{V}_t$ to calculate cross attention: 
\begin{equation}
\mathrm{Attn}(\boldsymbol{Q}_t, \boldsymbol{K}_t, \boldsymbol{V}_t) = \boldsymbol{\alpha}_t \boldsymbol{V}_t,\ 
\boldsymbol{\alpha}_t = \mathrm{Softmax} \left( \frac{\boldsymbol{Q}_t \boldsymbol{K}_t^T}{\sqrt{d'}} \right).
\end{equation}
Here, $\boldsymbol{\alpha}_t \in \mathcal{R}^{B \times N \times N}$ represents the historical attention of each node towards all nodes.

In this approach, $\mathrm{Attn}(\boldsymbol{Q}_t, \boldsymbol{K}_t, \boldsymbol{V}_t)$ extracts and fuses global node representation from previous time step based on historical attention. Subsequently, $\mathrm{Attn}(\boldsymbol{Q}_t, \boldsymbol{K}_t, \boldsymbol{V}_t)$ is transformed to $d_s$ dimensions through an MLP layer, and residual connections are applied at each layer using the current time step's node representation $\boldsymbol{N}_{t}$. This yields the historically enhanced node representation $\boldsymbol{N}_{t}^{S_*}$.

Note that while applying Dropout in linear layers is a common regularization technique for attention mechanisms, using standard Dropout between time steps can be harmful for RNN-based models~\cite{zaremba2014recurrent}. For RNN-based temporal models, introducing continuous noise is a more reliable approach. Hence, Variational Dropout~\cite{kingma2015variational} is adopted in the MLP layers to replace standard Dropout.

For an encoder or a decoder that is of $T$ time steps, at time step $t$, the above process is formalized as follows to obtain spatial correlation enhanced node representation $\boldsymbol{N}_{t}^{S_*}$:
\begin{equation}
    \boldsymbol{N}_{0} := \boldsymbol{N},
\end{equation}
\begin{equation}
    \boldsymbol{N}_{t}^{\mathrm{S}_*} = \mathrm{SCE}_{*}(\boldsymbol{N}_{t}, \boldsymbol{N}_{t-1}), \forall t = 1, \ldots, T.
\label{eq:sce}
\end{equation}

\subsubsection{Temporal Correlation Enhancement (TCE).}

We further enhance the node representation based on temporal correlations. Since SCE enables each node to extract historical information from all nodes, TCE allows each node to fuse representations from its previous time step. In this way, when the node representation is used to generate meta-parameters and an adjacency matrix, TCE can help mitigate abrupt changes between time steps and enhance temporal smoothness. Specifically, inspired by the process of updating hidden states using the update gate in GRU, MetaDG further leverages update gate $\boldsymbol{z}_{t - 1}$ to update the dynamic node representation $\boldsymbol{N}_{t}$ to be $\boldsymbol{N}_{t}^{\mathrm{T}_*}$: 
\begin{equation}
    \boldsymbol{\hat{z}}_{t - 1} = \mathrm{sigmoid}(\mathrm{FC}_z(\boldsymbol{z}_{t - 1})),
\end{equation}
\begin{equation}
    \boldsymbol{N}_{t}^{\mathrm{T}_*} = \boldsymbol{\hat{z}}_{t - 1} \odot \boldsymbol{N}_{t - 1} + (1 - \boldsymbol{\hat{z}}_{t - 1}) \odot \boldsymbol{N}_{t}.
\end{equation}
Here, $\boldsymbol{z}_{t - 1} \in \mathcal{R}^{B \times N \times d_H}$ is the update gate in GRU at time step $t-1$. In other words, constructing temporal correlations for node representations across time steps mimics the mechanism where the update gate in GRU associates hidden states across consecutive time steps.

For an encoder or decoder that is of $T$ time steps, at time step $t$, the above process is formalized as follows to obtain temporal correlation enhanced node representation $\boldsymbol{N}_{t}^{T_*}$:
\begin{equation}
\boldsymbol{N}_{t}^{T_*} = \mathrm{TCE}_{*}(\boldsymbol{N}_{t}, \boldsymbol{N}_{t-1}), \forall t = 2, \ldots, T.
\label{eq:tce}
\end{equation}

\subsubsection{Spatio-Temporal Correlation Enhancement (STCE).}

To simultaneously establish cross-time-step node correlations across both spatial and temporal dimensions, $\mathrm{SCE}_{*}(\cdot)$ given by Equation \ref{eq:sce} and $\mathrm{TCE}_{*}(\cdot)$ given by Equation \ref{eq:tce} are connected in series, yielding the enhanced representation $\boldsymbol{N}_{t}^{*}$ based on spatio-temporal correlations:
\begin{equation}
\boldsymbol{N}_{t}^{*} = \mathrm{STCE}_{*}(t), \forall t = 1, \ldots, T.
\label{eq:stce-node}
\end{equation}
where
\begin{equation}
\mathrm{STCE}_{*}(t) := 
\begin{cases}
    \mathrm{SCE}_{*}(\boldsymbol{N}_{t}, \boldsymbol{N}_{t-1}) & t=1, \\
    \mathrm{TCE}_{*}(\mathrm{SCE}_{*}(\boldsymbol{N}_{t}, \boldsymbol{N}_{t-1}), \boldsymbol{N}_{t-1}^{*}) & else.
\end{cases}
\label{eq:stce}
\end{equation}

$\mathrm{SCE}_{*}(\cdot)$ can extract information from global historical node representations, while $\mathrm{TCE}_{*}(\cdot)$ can smooth the differences across time steps. To obtain effective enhancement of node representations, we choose to fuse before smooth, that is $\mathrm{SCE}_{*}(\cdot)$ before $\mathrm{TCE}_{*}(\cdot)$, to eventually get $\mathrm{STCE}_{*}(\cdot)$.

\subsection{Dynamic Graph Qualification}

In this subsection, we propose the dynamic graph qualification (DGQ) module, which will adjust edge weights of the dynamic adjacency matrix based on the reliability of information propagation. The increase in the qualification of information propagation on edges can make graph convolution more effective \cite{ungsl}. This inspires us that qualifying propagated information may also be useful for graph convolution of GCRU at each time step. Indeed, for graph convolution in GCRU, the information propagated on the graph includes both of the current step input and the previous step hidden state, and the current step output will in turn, be used to construct the current step hidden state. That is, both current and historical information are propagated on the graph in GCRU. Since the recurrent nature of GRU may lead to error accumulation, qualification of graph convolution may be even important for GCRU than GCN. The idea is to adjust edge weights based on current and previous enhanced node representations. For edges that are reliable in current-history interactions, edge weights will be strengthened; otherwise, edge weights will be weakened. In this module, we aim to obtain an edge-weight adjustment matrix $\boldsymbol{\phi}_t$, which will refine the raw dynamic graph $\boldsymbol{A}_t$ to be $\tilde{\boldsymbol{A}}_t$. $\boldsymbol{A}_t$ will be given out in the next subsection.

Specifically, to generate such a \textbf{mask} $\boldsymbol{\phi}_t \in R^{B \times N \times N}$ that can adjust edge weights, we use STCE to generate an enhanced node representation, denoted as $\boldsymbol{N}_{t}^{m} = \mathrm{STCE}_{m}(t)$, where $\mathrm{STCE}_{*}(t)$ is given by Equation \ref{eq:stce}. We then use $\boldsymbol{N}_{t}^{m}$ and $\boldsymbol{N}_{t-1}^{m}$ to measure the qualification of edges based on cross-time-step similarities, denoted as $\boldsymbol{P}_t$:
\begin{equation}
    \boldsymbol{P}_t = \mathrm{asym}(ReLU(\boldsymbol{M} \odot (\boldsymbol{N}_{t}^{m} \cdot {\boldsymbol{N}_{t-1}^{m}}^T))),
\end{equation}
where $\boldsymbol{M} = (m_{ij})_{N \times N}$ denotes the static 0-1 adjacency matrix calculated by the inner product of $\boldsymbol{N}$, i.e., let $A = N\cdot N^T$, if $a_{ij}>0$ (i.e., $e_{ij} \in E$), then $m_{ij}=1$; else, $m_{ij}=0$. $\mathrm{asym}(\cdot)$ denotes row normalization. The adoption of $\boldsymbol{M}$ can limit the dynamics within a range, hence only edges that satisfy $e_{ij} \in E$ will be strengthened.

Based on the edge qualification matrix $\boldsymbol{P}_t$, we further need to decide which edge to strengthen or weaken. To do this, we first calculate the node-wise threshold $\boldsymbol{\epsilon}_{t} \in \mathcal{R}^{N \times 1}$ as the criterion:
\begin{equation}
    \boldsymbol{\epsilon}_{t;\ i} = \boldsymbol{P}_{t;\ (i, i)} \ \sigma(\boldsymbol{N}_{{t;\ i}}^{m} \cdot \boldsymbol{\epsilon}),\ \forall v_i \in V.
\end{equation}
Here, $\boldsymbol{\epsilon}  \in \mathcal{R}^{d_s \times 1}$ is the threshold pool, $\sigma(\cdot)$ denotes sigmoid activation. We use $\boldsymbol{P}_{t;\ (i, i)}$ as the threshold baseline to ensure that for $\forall v_i \in V$, $e_{ii}$ will not be weakened, which will stabilize the training process.

We follow the strategy of ``proportional strengthen and fixed weaken'' given in UnGSL \cite{ungsl} to do edge-weight adjustment. However, considering the complexity of the dynamic graph, we no longer use fixed coefficients. Instead of that, we calculate adaptive scaling coefficients $\boldsymbol{\beta}_{t}$. The non-zero elements of positive mask $\boldsymbol{M}_t^{pos}$ and negative mask $\boldsymbol{M}_t^{neg}$ are to be strengthened and weakened according to $\boldsymbol{\beta}_{t}$ to obtain edge-weight adjustment matrix $\boldsymbol{\phi}_t$:
\begin{equation}
    \boldsymbol{\phi}_t = \boldsymbol{\beta}_{t} \odot \boldsymbol{M}_t^{pos} + \boldsymbol{\beta}_{t} \odot \boldsymbol{M}_t^{neg},
\end{equation}
where
\begin{equation}
\boldsymbol{M}_t^{pos} = 
\begin{cases}
    \sigma(\boldsymbol{P}_{t;\ (i, j)} - \boldsymbol{\epsilon}_{t;\ (i, i)}) & \mathrm{if}\ \boldsymbol{P}_{t;\ (i, j)} - \boldsymbol{\epsilon}_{t;\ (i, i)} \geq 0, \\
    0 & \mathrm{else}.
\end{cases}
\end{equation}
\begin{equation}
    \boldsymbol{M}_t^{neg} = 
\begin{cases}
    0 & \mathrm{if}\ \boldsymbol{P}_{t;\ (i, j)} - \boldsymbol{\epsilon}_{t;\ (i, i)} \geq 0, \\
    1 & \mathrm{else}.
\end{cases}
\end{equation}
\begin{equation}
    \boldsymbol{\beta}_{t} = \mathrm{exp}(\mathrm{InstanceNorm}(\boldsymbol{M}_t^{pos}) \cdot \delta).
\end{equation}
Here, to obtain $\boldsymbol{M}_t^{pos}$ and $\boldsymbol{M}_t^{neg}$, we compare between edge qualification matrix $\boldsymbol{P}_t$ and the threshold $\boldsymbol{\epsilon}_{t}$. $\mathrm{InstanceNorm}(\cdot)$ is used to normalize the graph \cite{ulyanov2016instance}, so that the edges to be strengthened (weakened) will become positive (negative), and thus will be larger (smaller) than 1 after $\mathrm{exp}(\cdot)$ to serve as the scaler. $\delta$ is a scaler serves for effective exponential.

For an encoder or a decoder that is of $T$ time steps, at time step $t$, the above process is formalized as follows to obtain the edge-weight adjustment matrix $\boldsymbol{\phi}_t$ for further refinement of the dynamic adjacency matrix:
\begin{equation}
    \boldsymbol{N}_{0}^{m} := \boldsymbol{N}
\end{equation}
\begin{equation}
    \boldsymbol{\phi}_t = \mathcal{\varphi}(\boldsymbol{N}_{t}^{m}, \boldsymbol{N}_{t-1}^{m}), \forall t = 1, \ldots, T.
    \label{eq:phi}
\end{equation}

\subsection{MetaDG Convolutional Recurrent Unit}

GCRU \cite{agcrn, dgcrn, himnet} is the combination of GRU and GCN, which uses fixed parameters and a static graph. In this subsection, we propose Meta Dynamic Graph Convolutional Reurrent Unit (Meta-DGCRU), where for each time step $t$, we replace the parameters and graph used in graph convolution with meta-parameters $\boldsymbol{\theta}_t$ and dynamic graph $\tilde{\boldsymbol{A}_{t}}$ based on the previous modules. Specifically, for each time step $t$, the standard GCRU is replaced by the following:
\begin{equation}
    \boldsymbol{z}_t = \sigma({\Theta}_{z \star \mathcal{G}}^t [\boldsymbol{X}_t \Vert \boldsymbol{H}_{t-1}]),
    \label{eq:zgt}
\end{equation}
\begin{equation}
    \boldsymbol{r}_t = \sigma({\Theta}_{r \star \mathcal{G}}^t [\boldsymbol{X}_t \Vert \boldsymbol{H}_{t-1}]),
    \label{eq:rgt}
\end{equation}
\begin{equation}
    \boldsymbol{c}_t = \sigma({\Theta}_{c \star \mathcal{G}}^t [\boldsymbol{X}_t \Vert \boldsymbol{r}_t \odot \boldsymbol{H}_{t-1}]),
    \label{eq:cgt}
\end{equation}
\begin{equation}
    \boldsymbol{H}_{t} = \boldsymbol{z}_t \odot \boldsymbol{H}_{t-1} + (1 - \boldsymbol{z}_t) \odot \boldsymbol{c}_t.
    \label{eq:hgt}
\end{equation}
Here, $\boldsymbol{X}_t$ denotes the input which concatenates traffic flow and time, $\boldsymbol{H}_t$ denotes the output of hidden state, $\boldsymbol{z}_t$ and $\boldsymbol{r}_t$ denote update gate and reset gate, ${\Theta}_{z \star \mathcal{G}}^t (\cdot)$, ${\Theta}_{r \star \mathcal{G}}^t (\cdot)$, and ${\Theta}_{c \star \mathcal{G}}^t (\cdot)$ denote the 1-hop graph convolution that use the generated meta-parameters (represented by $\boldsymbol{\theta}_t$) and dynamic graph $\tilde{\boldsymbol{A}_{t}}$.

To dynamically generate meta-parameters, raw adjacency matrix and edge-weight adjustment matrix, we first generate raw dynamic node embedding $\boldsymbol{N}_t$, and do spatio-temporal correlation enhancement based on $\boldsymbol{N}_t$ to get $\boldsymbol{N}_{t}^{p}, \boldsymbol{N}_{t}^{g}, \boldsymbol{N}_{t}^{m} \in R^{B \times N \times d_s}$ as given by Equation \ref{eq:stce-node} where we substitute $*$ with $p, g, m$. Here, $\boldsymbol{N}_{t}^{p}$ will be used to generate meta-parameters $\boldsymbol{\theta}_t$, $\boldsymbol{N}_{t}^{g}$ will be used to generate raw adjacency matrix $\boldsymbol{A}_{t}$, and $\boldsymbol{N}_{t}^{m}$ will be used to generate edge-weight adjustment matrix $\boldsymbol{\phi}_t$:
\begin{itemize}[]
    \item For parameter pool $\boldsymbol{\Theta} \in R^{d_s \times I \times O}$, use $\boldsymbol{N}_{t}^{p}$ to generate node-wise \textbf{meta-parameter $\boldsymbol{\theta}_t \in R^{B \times N \times I \times O}$}:
    \begin{equation}
    \boldsymbol{\theta}_t = \boldsymbol{N}_{t}^p \boldsymbol{\Theta}.
    \end{equation}

    \item Use $\boldsymbol{N}_{t}^{g}$ to calculate similarities for node pairs, to generate \textbf{raw adjacency matrix $\boldsymbol{A}_{t} \in R^{B \times N \times N}$}:
    \begin{equation}
    \boldsymbol{A}_{t} = \mathrm{ReLU}(\boldsymbol{N}_{t}^{g} \cdot {\boldsymbol{N}_{t}^{g}}^T).
    \label{eq:adj-t}
    \end{equation}

    \item Use $\boldsymbol{N}_{t}^{m}$ to generate \textbf{edge-weight adjustment matrix $\boldsymbol{\phi}_t$} which is given by Equation \ref{eq:phi}.

\end{itemize}

Hence, we can eventually get the adjacency matrix $\tilde{\boldsymbol{A}_{t}}$ of the dynamic graph based on row normalization:
\begin{equation}
    \tilde{\boldsymbol{A}_{t}} = \mathrm{asym}(\boldsymbol{\phi}_t \odot \boldsymbol{A}_{t}).
\end{equation}

Moreover, as node representation is often set to be a low-dimensional vector, the raw adjacency matrix generated by $\boldsymbol{N}_{t}^{g}$ will also be of low rank. To fix this problem, we can incorporate continuous time $\boldsymbol{T}^c_t = \sqrt{\frac{1}{d}} [\mathrm{cos}(\omega_1 \tau_t), \mathrm{sin}(\omega_1 \tau_t), \ldots, \mathrm{cos}(\omega_{d_c} \tau_t), \mathrm{sin}(\omega_{d_c} \tau_t)]$ \cite{xu2020inductive} to generate a high-dimensional node representation $\boldsymbol{N}_{t}^{h} = \boldsymbol{T}^c_t \mathrm{FC}_h(\boldsymbol{N}_{t}^{g})$. Hence, we can further update $\boldsymbol{A}_{t}$ given by Equation \ref{eq:adj-t} as follows:
\begin{equation}
    \boldsymbol{A}_{t} = \mathrm{ReLU}(\boldsymbol{N}_{t}^{h} \cdot {\boldsymbol{N}_{t}^{h}}^T) \odot \boldsymbol{A}_{t}.
\end{equation}

\section{Experiments}

\subsection{Experimental Setup}

\sstitle{Datasets}
We conduct experiments on four real-world traffic flow datasets, i.e., PEMS03, PEMS04, PEMS07, and PEMS08 \cite{stsgcn}. The time interval is 5 minutes. More detailed statistics of the datasets are shown in Table~\ref{tab:meta-data}. To ensure consistent data scales and enhance training stability, we perform Z-score normalization on raw inputs during data preprocessing ~\cite{dcrnn}. For fair comparison with prior works, we adopt the commonly-used dataset split from existing literature ~\cite{himnet}, partitioning data into training/validation/test sets at a 6:2:2 ratio.

\begin{table}[t]
\centering
\begin{tabular}{
    >{\centering\arraybackslash}p{14mm}|
    >{\centering\arraybackslash}p{12mm}
    >{\centering\arraybackslash}p{15mm}
    >{\centering\arraybackslash}p{25mm}}
\toprule
\textbf{Dataset}       & \#Sensors      & \#Timesteps & Time Range   \\
\midrule
\textbf{PEMS03}        & 358        & 26,185 & 09/2018-11/2018        \\ 
\textbf{PEMS04}        & 307        & 16,992 & 01/2018-02/2018        \\ 
\textbf{PEMS07}        & 883        & 28,224 & 05/2017-08/2017        \\ 
\textbf{PEMS08}        & 170        & 17,856 & 07/2016-08/2016       \\
\bottomrule
\end{tabular}
\caption{Dataset Descriptions.}
\label{tab:meta-data}
\end{table}

\sstitle{Baselines}
Baselines include typical models commonly used for spatio-temporal flow predictions, such as STGCN~\cite{stgcn}, DCRNN~\cite{dcrnn}, GWNet~\cite{gwn}, AGCRN~\cite{agcrn}, STSGCN~\cite{stsgcn}, STID~\cite{stid}, PDFormer~\cite{pdformer}, MegaCRN~\cite{megacrn}, DGCRN~\cite{dgcrn}, HimNet~\cite{himnet}, and ST-SSDL~\cite{st-ssdl}. Specifically, AGCRN, MegaCRN, and HimNet are meta-learning methods; while STSGCN, PDFormer, and DGCRN are dynamic methods.

\sstitle{Evaluation Metrics}
We use Mean Absolute Error (MAE), Root Mean Squared Error (RMSE), and Mean Absolute Percentage Error (MAPE, \%) to evaluate the performance.

\sstitle{Implementation}
Our method is implemented with PyTorch. Experiments are conducted on a workstation with one GeForce RTX 4090. 
We set time steps $T$ and $T'$ to be 12. For embedding dimensions $d_s$, $d_{tod}$, $d_{dow}$, and $d_c$, we set 12, 8, 8, 8 for PEMS03, 16, 12, 4, 6 for PEMS04, 16, 8, 8, 8 for PEMS07, and 12, 10, 2, 8 for PEMS08. The dimension $d_H$ of the hidden state is set to be 64. The dimension $d'$ of $\boldsymbol{Q}$, $\boldsymbol{K}$, and $\boldsymbol{V}$ in cross-attention of SCE is set to be 64. Batch size is set to be 8 for PEMS07, and 16 for others. $\delta$ is set to be 2. We train the model within 200 epochs, and will achieve early stop if validation loss has not been decreasing for 20 epochs. We use Huber Loss~\cite{huber1992robust} as the loss function.

\begin{table*}[t]
    \centering
    \begin{adjustbox}{width=0.85\textwidth, keepaspectratio}
    \begin{tabular}{
    c|c|ccc|ccc|ccc|ccc
    }
        \toprule
         \multicolumn{2}{c|}{Datasets} & \multicolumn{3}{c|}{PEMS03}      & \multicolumn{3}{c|}{PEMS04}    & \multicolumn{3}{c|}{PEMS07}    & \multicolumn{3}{c}{PEMS08}    \\ \midrule
         \multicolumn{2}{c|}{Metrics} & MAE     & RMSE      & MAPE      & MAE     & RMSE      & MAPE    & MAE     & RMSE      & MAPE      & MAE     & RMSE      & MAPE    \\ \midrule
         \multirow{11}{*}{Baselines} & STGCN & 15.91 & 27.46 & 16.09 & 19.64 & 31.49 & 13.45 & 21.89 & 35.38 & 9.28 & 16.09 & 25.42 & 10.57 \\
         & DCRNN & 15.63 & 27.26 & 15.89 & 19.63 & 31.28 & 13.57 & 21.31 & 34.28 & 9.15 & 15.23 & 24.25 & 10.28 \\
         & GWNet & 14.62 & \cellcolor{gray!20}{25.28} & 15.53 & 18.54 & 29.96 & 12.87 & 20.53 & 33.49 & 8.63 & 14.41 & 23.37 & 9.21 \\
         & AGCRN & 15.36 & 26.73 & 15.87 & 19.34 & 31.23 & 13.37 & 20.57 & 34.21 & 8.70 & 15.31 & 24.43 & 10.12 \\
         & STSGCN & 15.05 & 25.79 & 15.79 & 18.64 & 30.37 & 12.81 & 20.13 & 33.89 & 8.58 & 14.37 & 23.31 & 9.27 \\
         & STID & 15.33 & 27.40 & 16.40 & 18.38 & 29.95 & \cellcolor{gray!20}{12.04} & 19.61 & 32.79 & 8.30 & 14.21 & 23.28 & 9.27 \\
         & PDFormer & 14.92 & 25.43 & 15.77 & 18.32 & 30.02 & {12.07} & 19.88 & 32.89 & 8.53 & 13.64 & 23.44 & 9.24 \\
         & MegaCRN & 14.58 & 25.83 & \cellcolor{gray!20}{14.78} & 18.75 & 30.46 & 12.75 & 19.81 & 32.87 & 8.37 & 14.75 & 23.76 & 9.49 \\
         & DGCRN & 14.63 & 25.74 & 14.99 & 19.09 & 31.48 & 12.57 & 19.87 & 32.91 & 8.46 & 14.59 & 23.57 & 9.44 \\
         & HimNet & 15.14 & 26.78 & 15.55 & 18.31 & 30.15 & 12.18 & 19.50 & 32.79 & 8.29 & \cellcolor{gray!20}{13.57} & 23.25 & \cellcolor{gray!20}{8.99} \\
         & ST-SSDL & \cellcolor{gray!20}{14.56} & 25.79 & 15.08 & \cellcolor{gray!20}{18.13} & \cellcolor{gray!20}{29.77} & 12.57 & \cellcolor{gray!20}{19.24} & \cellcolor{gray!20}{32.77} & \cellcolor{gray!20}{8.10} & 13.88 & \cellcolor{gray!20}{23.15} & 9.08 \\
         \midrule
         \textbf{ours} & \textbf{MetaDG}     & \cellcolor{gray!60}\textbf{14.29} & \cellcolor{gray!60}\textbf{24.93} & \cellcolor{gray!60}{14.64} & \cellcolor{gray!60}\textbf{17.80} & \cellcolor{gray!60}\textbf{29.46} & \cellcolor{gray!60}\textbf{11.70} & \cellcolor{gray!60}\textbf{18.79} & \cellcolor{gray!60}{32.29} & \cellcolor{gray!60}\textbf{7.89} & \cellcolor{gray!60}\textbf{13.04} & \cellcolor{gray!60}{22.53} & \cellcolor{gray!60}{8.58}  \\ \midrule
         \multirow{6}{*}{Ablations} & w/o SCE      & 14.88 & 25.79 & 15.33 & 18.20 & 30.60 & 11.96 & 19.39 & 33.69 & 8.17 & 13.33 & 22.88 & 8.81  \\ 
         & w/o TCE      & 14.35 & 25.38 & \underline{14.50} & \underline{17.87} & \textbf{29.46} & \underline{11.75} & 18.95 & \underline{32.18} & 8.07 & \underline{13.06} & \underline{22.48} & 8.60  \\ 
         & w/o STCE      & 14.98 & 26.21 & 15.20 & 18.17 & 30.35 & 11.94 & 19.28 & 33.45 & 8.08 & 13.37 & 22.97 & 8.77  \\ 
         & w/o DGQ      & 14.48 & \underline{25.11} & 14.80 & 17.88 & \underline{29.54} & 11.80 & \underline{18.91} & 32.52 & \underline{7.90} & \underline{13.06} & 22.54 & \underline{8.56}  \\ 
         & TSCE      & \underline{14.33} & 25.18 & \textbf{14.40} & 17.92 & 29.74 & 11.80 & \underline{18.91} & \textbf{32.07} & 8.08 & \textbf{13.04} & 22.55 & \textbf{8.55}  \\ 
         & Joined      & 14.55 & 26.32 & 14.90 & 18.00 & 29.85 & 11.88 & 18.93 & 32.44 & 7.98 & \textbf{13.04} & \textbf{22.47} & 8.57  \\
        \bottomrule
    \end{tabular}
    \end{adjustbox}
    \caption{Overall Performance and Ablation Study. For overall performance, use \colorbox{gray!60}{dark gray} and \colorbox{gray!30}{light gray} to mark the best and second best separately; for ablation study, use \textbf{bold} and \underline{underline} to mark the best and second best separately.}
    \label{tab:metadg-performance-and-ablation}
\end{table*}

\subsection{Overall Performance}

As shown in Table \ref{tab:metadg-performance-and-ablation}, comparison with baseline methods demonstrates that MetaDG achieves significantly superior results in traffic flow prediction by generating the dynamic node representation at each time step for producing node parameters and adjacency matrix. AGCRN, MegaCRN, and HimNet are GCRU-based meta-learning methods. Compared with these static meta-learning methods, MetaDG achieves better results by dynamically generating model components for each time step. STSGCN, PDFormer, and DGCRN are dynamic methods that take the dynamics of spatial topology into consideration. In comparison, MetaDG considers dynamics as a more intrinsic nature by using dynamic graph structure to model more intermediates, including meta-parameters, raw adjacency matrix, and edge-weight adjustment matrix. By extending the usage of dynamics into a more inherent and broader level, we push the modeling of spatio-temporal correlations and heterogeneities towards ST-unification, which has been tested to be effective.

\subsection{Ablation Study}

We further compare the performance of MetaDG with six variants to prove the effectiveness of each module:
\begin{itemize}[leftmargin=*]
\item  \textbf{MetaDG-w/o SCE}, which removes spatial correlation enhancement of dynamic node representations.
\item \textbf{MetaDG-w/o TCE}, which removes temporal correlation enhancement of dynamic node representations.
\item \textbf{MetaDG-w/o STCE}, which removes spatio-temporal correlation enhancement of dynamic node representations.
\item \textbf{MetaDG-w/o DGQ}, which removes dynamic graph qualification, so that the dynamic graph structures are not refined based on message-passing reliability.
\item \textbf{MetaDG-TSCE}, which switches the enhancement order of SCE and TCE, i.e., smoothing-before-fusion.
\item \textbf{MetaDG-Joined}, which substitues $\boldsymbol{N}_t^p$, $\boldsymbol{N}_t^g$, and $\boldsymbol{N}_t^m$ with a joined dynamic node embedding $\boldsymbol{N}_t^{joined}$.
\end{itemize}

As shown in Table~\ref{tab:metadg-performance-and-ablation}, removing SCE, TCE, or STCE degrades model performance, demonstrating the critical role of spatio-temporal correlations in optimizing dynamic node representations for meta-parameters and adjacency matrix generation. Removing DGQ similarly reduces effectiveness, highlighting the necessity of refinement of adjacency matrices based on estimation of message-passing reliability in a GCRU-based model. MetaDG-TSCE reverses the execution order of SCE and TCE to smoothing-before-fusion, which also leads to performance deterioration in most of the metrics. This validates the rationality of the enhancement order of fusion-before-smoothing. MetaDG-Joined shows that in most cases, different model components (i.e., $\boldsymbol{\theta}_t$, $\boldsymbol{A}_t$, $\boldsymbol{\phi}_t$) may need different kinds of correlations, and thus it is reasonable to generate different model components using separately enhanced node representations.

\subsection{Hyperparameter Study}

\begin{figure}[b]
\centering
\includegraphics[width=0.98\columnwidth]{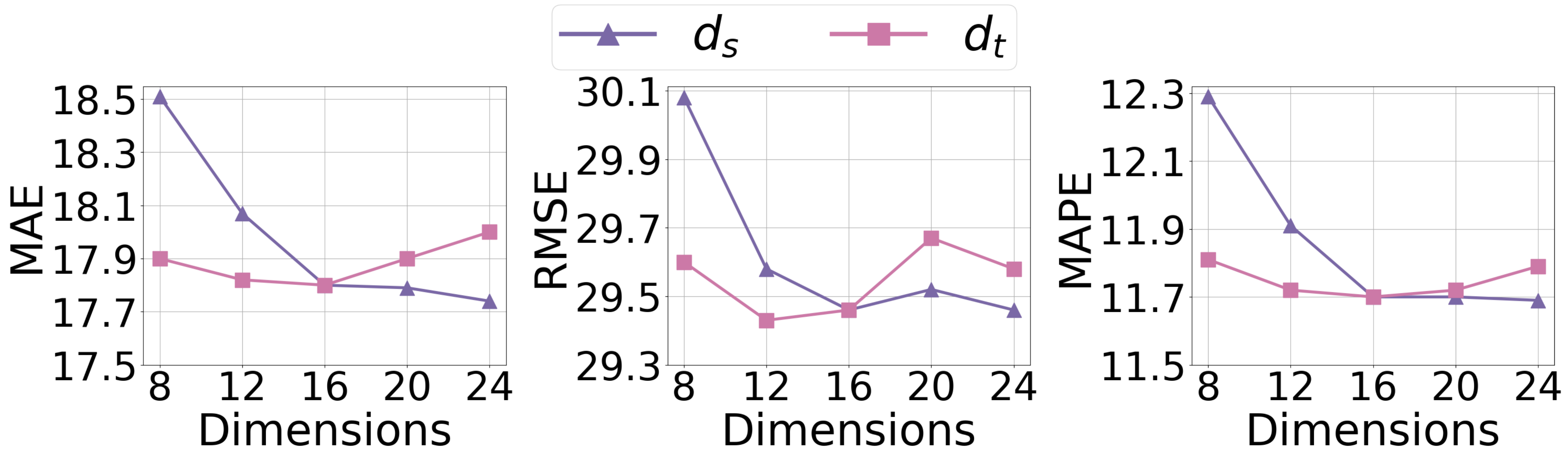}
\caption{Hyperparameter Study.}
\label{fig:metadg-hyperparam}
\end{figure}

Figure \ref{fig:metadg-hyperparam} shows the performance of different dimensions of embedding vectors $d_s$ and $d_t$, taking PEMS04 dataset as an example. The performances of different embedding dimensions always exhibit relatively sharp inflection points \cite{dgcrn, himnet}. Yet for MetaDG, after the node dimension $d_s$ reaches $16$, increasing $d_s$ slightly further improves the performance. Time dimension $d_t$ still reaches an inflection point at $d_t = 16$, yet the increase of the criterions on both sides of the inflection point is relatively slight. By learning dynamic graph structure, MetaDG significantly enhances the organizational capability of embeddings, thus reduce the effort for finding effective hyperparameters.

\subsection{Performance w.r.t. Different Time Steps}

\begin{figure}[htb]
\centering
\includegraphics[width=0.98\columnwidth]{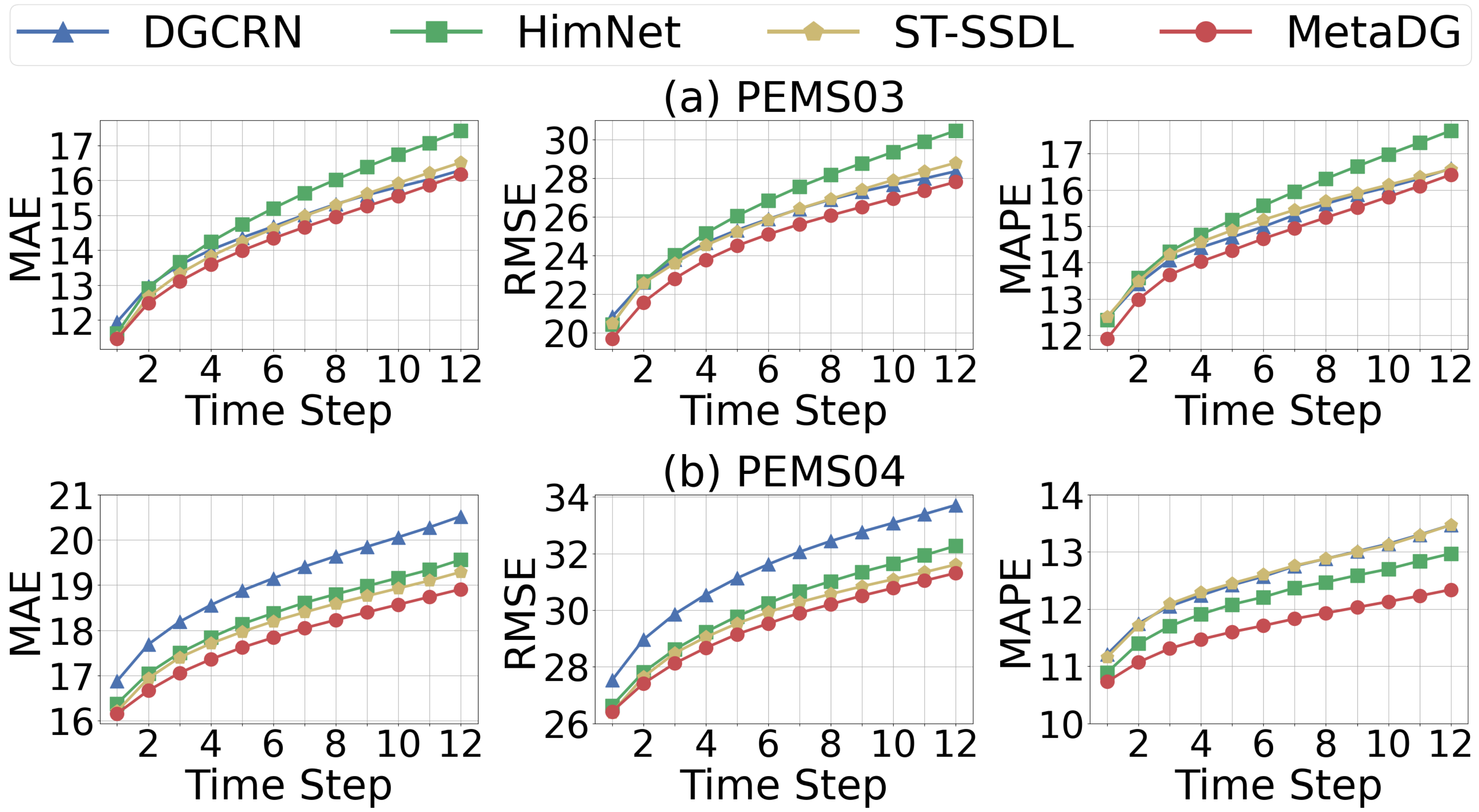}
\caption{Per Time Step Performance.}
\label{fig:metadg-step}
\end{figure}

To better show the effect of adopting dynamic to enhance ST-unification, in Figure \ref{fig:metadg-step}, we compare per time step performance of typical methods on PEMS03/04. We can see that MetaDG has more advantage in long-term predictions.

\subsection{Efficiency Comparison}

In Table~\ref{tab:effeciency}, we compare the computational efficiency of MetaDG with 3 typical baseline models and the variation of MetaDG-Joined on PEMS03. The result shows that MetaDG reduces time compared to typical dynamic method DGCRN and reduces parameters compared to typical meta-learning method HimNet. Moreover, MetaDG-Joined can achieve a comparable inference time with ST-SSDL.

\begin{table}[h]
\begin{adjustbox}{width=0.96\columnwidth, totalheight=0.96\columnwidth, keepaspectratio}
\centering
\begin{tabular}{
    c|ccccc}
\toprule
\textbf{PEMS03}       & DGCRN      & HimNet & ST-SSDL & MetaDG & MetaDG-Joined   \\
\midrule
\textbf{\#Params}       & 208K      & 2742K & 234K & 666K & 649K        \\ 
\textbf{Train}       & 287s      & 175s & 172s & 250s & 197s        \\ 
\textbf{Infer}       & 33s      & 19s & 19s & 23s & 20s        \\
\bottomrule
\end{tabular}
\end{adjustbox}
\caption{Efficiency Comparison.}
\label{tab:effeciency}
\end{table}

\section{Related Work}

\subsubsection{Classical Spatio-temporal Prediction Methods.}
Deep learning based spatio-temporal prediction models often model temporal and spatial dimensions separately using different structures. For the temporal dimension, RNNs \cite{gru_conference, lstm_paper, agcrn} and CNNs \cite{lenet5, stgcn, gwn} are always used, while for the spatial dimension, GNNs, e.g., GCN \cite{gcn, stgcn, gwn} and GAT~\cite{gat, pan2019urban}, are always selected. 
MLP-based spatio-temporal modeling is also reported in recent studies~\cite{WANG2024111463}.
However, the ST-isolated nature may hinder the modeling effect.

\subsubsection{Meta-Learning-based Spatio-temporal Prediction Methods.} Considering spatio-temporal heterogeneities is a useful strategy, using meta-learning is a common choice~\cite{ruan2022service,wang2024spatial}. Specifically, AGCRN \cite{agcrn} models spatial structures through static node representations to generate adaptive graphs and node-level parameters. MegaCRN \cite{megacrn} uses encoder outputs to characterize traffic patterns, and generates an adaptive graph for the decoder. HimNet \cite{himnet} employs adaptive adjacency matrices and node-level parameters in both encoder and decoder, leveraging spatial, temporal, and spatio-temporal embeddings for the generation of parameters and matrices. Nevertheless, these methods model heterogeneities in an ST-isolated manner, and fail to model inter-time-step dynamics.

\subsubsection{Dynamic Spatio-temporal Prediction Methods.} Recent studies also emphasize the effectiveness of replacing statics with dynamics. The motivation is that the real-time spatial topology is variable, making a static graph insufficient to characterize the real situation.
STSGCN \cite{stsgcn} uses a localized spatio-temporal graph, while PDFormer \cite{pdformer} uses self-attention and explicitly models delayed message passing. Here, dynamics are modeled in a less flexible way. In comparison, DGCRN \cite{dgcrn} generates a dynamic graph for each time step, but uses hyper-GCN and fails to consider spatio-temporal heterogeneities. Effective message passing relies on the modeling of dynamics, and thus it is reasonable to extend the usage of dynamics from adjacency matrices to a broader scope.
For example, the dynamic structure of spatio-temporal nodes can be modeled and used to generate adjacency matrices, meta-parameters, and other intermediates of the model. Dynamics can bridge the gap between spatial and temporal dimensions, and push the modeling of correlations and heterogeneities from ST-isolated towards ST-unification.

\section{Conclusion}

Traffic flow prediction is a typical spatio-temporal prediction problem. In this paper, we propose MetaDG, which is a GCRU-based model considering dynamics and heterogeneities simultaneously. Specifically, we generate dynamic node embedding and enhance it according to spatio-temporal correlations. We use the dynamic node representation to generate meta-parameters and a raw adjacency matrix. We further generate an edge-weight adjustment matrix by qualifying the reliability of message-passing, which will be used to refine the raw adjacency matrix. This effort not only brings the base model structure but also the modeling of heterogeneities from ST-isolated towards ST-unification. For future work, we will try to extend the ST-unifying functionality of dynamics into a broader scale of base models and scenarios.

\section{Acknowledgments}
This research is supported by National Natural Science Foundation of China (No. 62306033, 42371480).

\bibliography{main}

\end{document}